\pdfoutput=1
\documentclass[11pt]{article}

\usepackage[T1]{fontenc}
\usepackage[utf8]{inputenc}
\usepackage[margin=1in]{geometry}
\usepackage{times}
\usepackage{amsmath,amssymb,amsfonts,amsthm,bm}
\usepackage{booktabs}
\usepackage{multirow}
\usepackage{subcaption}
\usepackage{wrapfig}
\usepackage{nicefrac}
\usepackage{microtype}
\usepackage{xcolor}
\usepackage{graphicx}
\usepackage{colortbl}
\usepackage{adjustbox}
\usepackage[round,authoryear]{natbib}
\usepackage{url}
\usepackage[colorlinks=true,allcolors=blue]{hyperref}

\usepackage{amsmath,amsfonts,bm}

\def\eqref#1{equation~\ref{#1}}

\def\1{\bm{1}}

\DeclareMathAlphabet{\mathsfit}{\encodingdefault}{\sfdefault}{m}{sl}
\SetMathAlphabet{\mathsfit}{bold}{\encodingdefault}{\sfdefault}{bx}{n}

\newtheorem{theorem}{Theorem}[section]

\newtheorem{remark}[theorem]{\textbf{Remark}}

\newtheorem{finding}[theorem]{\textbf{Finding}}

\makeatletter
\renewcommand{\eqref}[1]{\textup{\tagform@{\ref{#1}}}}
\makeatother

\title{\texorpdfstring{%
\makebox[0pt][r]{%
  \raisebox{-0.5\height}{\includegraphics[height=2.8\baselineskip]{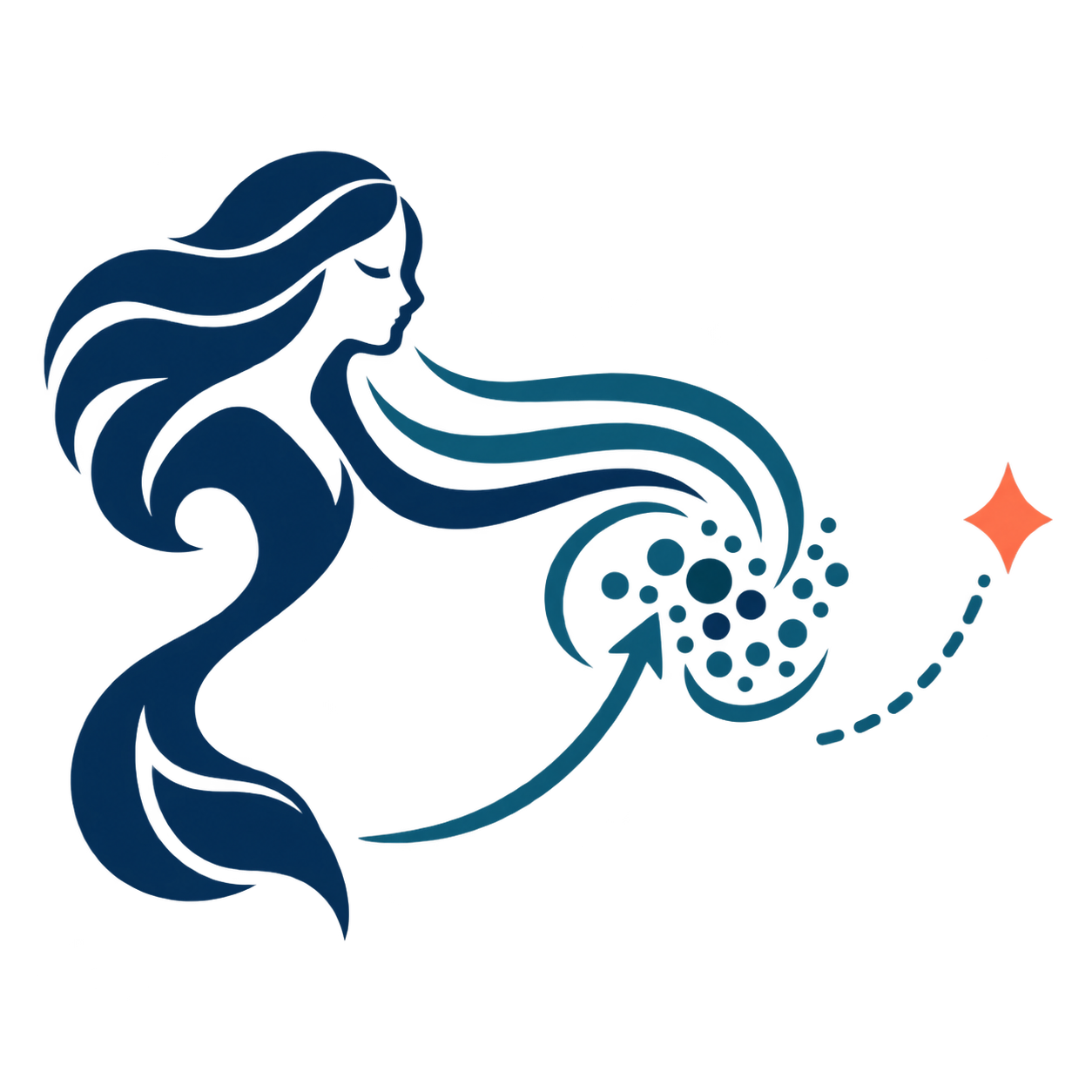}}%
  \hspace{0.1em}%
}%
\parbox[c]{0.74\textwidth}{\centering
The Sirens' Song: When Proximal Background Context Overshadows Distant Evidence}%
}{The Sirens' Song: When Proximal Background Context Overshadows Distant Evidence}}

\author{Xiaoyu Yang, Jie Lu, Wei Duan, En Yu \\
Australian Artificial Intelligence Institute (AAII), \\ 
Faulty of Engineering and Information Technology,\\
University of Technology Sydney, Australia.
}

\date{}

\begin{document}

\maketitle

\begin{abstract}
Long-context LLMs focus on retrieving distant evidence from extensive context, yet existing work has largely focused on overcoming distance alone.
In this work, we identify the Proximity Trap, insufficient attention to distant evidence often arises less from distance itself than from cumulative competition with abundant, task-irrelevant proximal background. 
To address the Proximity Trap, we introduce LYRA (\textbf{L}ong-context heav\textbf{Y}-tailed \textbf{R}elevance \textbf{A}lignment), a t-distributed directional matching mechanism that reshapes the context retrieval distribution, directing more attention mass toward task-relevant evidence, while preserving the relative positional information encoded.
Extensive experiments on LongBench-v2, RULER, and LongBench demonstrate consistent improvements across context lengths and task categories. We further introduce ProxBench, a multi-level fine-grained benchmark for evaluating distant evidence utilization under increasing proximal background interference. 

\vspace{2mm}
\centering{Project page: \url{https://xiaoyuyoung.github.io/LYRA/}}

\end{abstract}

\section{Introduction}

Recent work on long-context LLMs  \citep{wang2026proxy,zhang2026s2o,mudarisov2025limitations} has focused on improving access to distant evidence. However, access alone does not ensure that such evidence will be used, while proximal, task-irrelevant context may also compete for attention. This raises a less examined question: when a model overlooks distant evidence, is distance alone responsible, or does proximal background also impede its use?

Prioritizing recent context is a reasonable inductive bias, as nearby tokens often provide useful information for the current prediction. Accordingly, modern LLMs commonly encode relative positions through RoPE \citep{su2024roformer}, making QK scores explicitly dependent on relative distance and potentially attenuating the scores of distant tokens. While this positional bias supports local context modeling, positional proximity does not necessarily indicate task relevance: essential evidence may lie far from the query, whereas nearby content may provide little information for the required prediction. This mismatch raises uncertainty about how attention is allocated when distant evidence competes with proximal background.

\begin{wrapfigure}{r}{0.35\textwidth}
    \centering
    \includegraphics[width=0.35\textwidth]{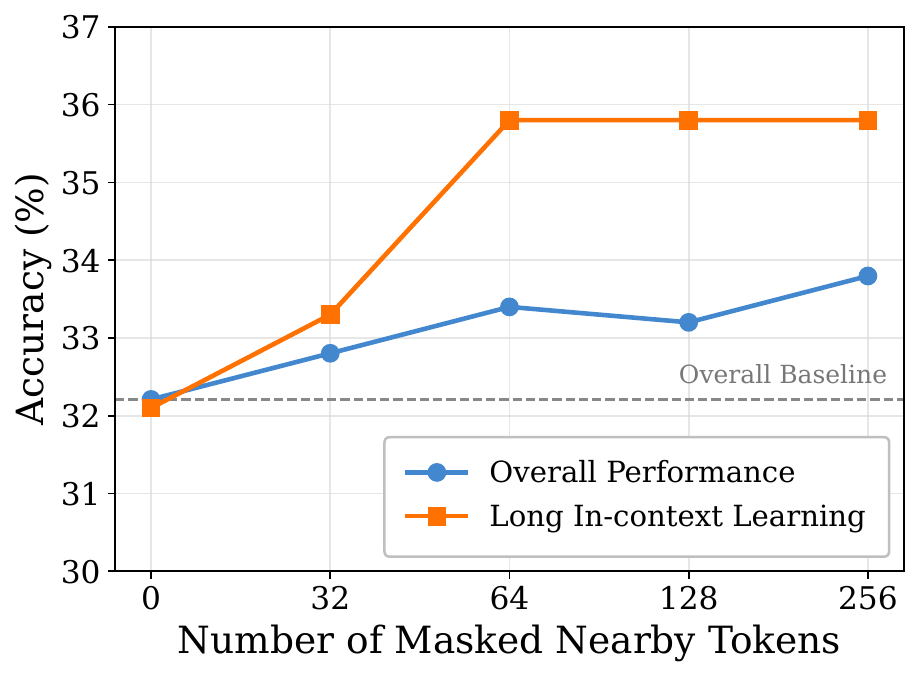}
    \caption{\textbf{Less attention to proximal background improves long-context understanding.} On LongBench-v2 \citep{bai2025longbench}, we progressively mask the context tokens nearest to the query and report accuracy over the full benchmark and the Long In-context Learning subset. The horizontal dashed line denotes the performance without masking, corresponding to zero masked tokens. }
    \label{fig:intro}
\end{wrapfigure}

In this work, we present a counter-intuitive observation: reducing attention to proximal background can improve long-context understanding. As shown in Fig.~\ref{fig:intro}, masking proximal background tokens consistently improves the overall accuracy on LongBench-v2 compared with the unmasked baseline, with a more pronounced improvement on Long In-context Learning. This result is striking because removing accessible context improves performance rather than degrading it. It suggests that the difficulty of using distant evidence cannot be attributed to distance alone; competition from proximal background also plays a critical role. We refer to this phenomenon as the \textit{Proximity Trap}.

This counter-intuitive observation is closely related to how attention combines positional distance with contextual competition. Long-context inputs are inherently non-uniform: the evidence required for a prediction is often sparse and distant, whereas much of the nearby context serves only as background. Relative positional encoding makes QK matching dependent on distance and can weaken the directional agreement between the query and distant evidence. Under softmax normalization, however, this weakened evidence competes not with a single nearby token, but with all proximal background tokens simultaneously. Although each background token may have limited task relevance, their contributions accumulate in the softmax denominator and can collectively draw attention away from distant evidence. This interaction between positional attenuation and cumulative background competition underlies the Proximity Trap and explains why masking proximal background can improve evidence utilization.

Consequently, we introduce LYRA, a simple t-distributed directional matching mechanism. Preserving distant evidence does not require explicitly favoring distant positions or manually suppressing proximal context. Instead, LYRA measures the directional agreement between the RoPE-transformed query and keys, and then applies a t-distributed transformation to reshape their score differences before softmax normalization. This design reduces misleading advantages held by weakly related proximal background while making strongly matched evidence more distinguishable. Crucially, LYRA replaces only the conventional QK scoring function, leaving RoPE, causal masking, softmax normalization, and value aggregation unchanged. As a result, LYRA preserves distant evidence according to relevance rather than position, allowing it to resist cumulative background competition without indiscriminately promoting all distant tokens.

In summary, our main contributions are as follows:
\begin{itemize}

    \item We characterize the Proximity Trap, a counter-intuitive phenomenon in which proximal, task-irrelevant background can collectively overshadow distant but relevant evidence.

    \item LYRA is introduced as a t-distributed directional matching mechanism that reshapes the context retrieval distribution to redistribute attention mass toward task-relevant evidence while preserving the relative positional information encoded.

    \item Extensive experiments are conducted on various benchmarks. The results demonstrate consistent improvements across different context lengths and task categories, supporting the effectiveness and generalization of LYRA.

    \item We introduce ProxBench, a multi-level controlled benchmark for evaluating distant evidence utilization under increasing proximal background interference.

\end{itemize}

\section{The Proximity Trap}
\label{sec:2}

In this section, we first introduce the preliminaries in Section ~\ref{sec:2.1}. Subsequently, we establish our central observation that proximal background context overshadows distant evidence in Section~\ref{sec:2.2}. Moreover, we provide an explanation of its underlying mechanism through cumulative attention competition in Section~\ref{sec:2.3}.

\subsection{Preliminaries}
\label{sec:2.1}

Given a query at the current decoding position $t$, let $\mathbf{q}_t \in \mathbb{R}^{d}$ denote the query representation and let $(\mathbf{k}_i,\mathbf{v}_i)$ denote the key--value pair at each historical position $i<t$. Standard attention first computes the QK score between the query and each key, followed by softmax normalization:
\begin{equation}
s_{t,i}=\frac{\mathbf{q}_t^{\top}\mathbf{k}_i}{\sqrt{d}},
\qquad
a_{t,i}=\frac{\exp(s_{t,i})}{\sum_{j<t}\exp(s_{t,j})}.
\label{eq:standard_attention}
\end{equation}
The resulting attention output is $\mathbf{o}_t=\sum_{i<t}a_{t,i}\mathbf{v}_i$. Since $a_{t,i}\geq 0$ and $\sum_{i<t}a_{t,i}=1$, all historical positions compete for a finite amount of attention weight. Our analysis therefore focuses on how the QK scores assigned to different historical positions determine this competition when the model generates the token at position $t$.

In particular, the QK score defined in Eq.~\eqref{eq:standard_attention} becomes explicitly position-dependent when RoPE \citep{su2024roformer} is applied to the query and key representations. Let $\mathbf{R}_m \in \mathbb{R}^{d \times d}$ denote the rotation matrix associated with position $m$. RoPE transforms the query and key as $\widetilde{\mathbf{q}}_t=\mathbf{R}_t\mathbf{q}_t$ and $\widetilde{\mathbf{k}}_i=\mathbf{R}_i\mathbf{k}_i$, respectively. The resulting QK score is:
\begin{equation}
s_{t,i}^{\mathrm{RoPE}}
=
\frac{\widetilde{\mathbf{q}}_t^{\top}\widetilde{\mathbf{k}}_i}{\sqrt{d}}
=
\frac{\mathbf{q}_t^{\top}\mathbf{R}_t^{\top}\mathbf{R}_i\mathbf{k}_i}{\sqrt{d}}
=
\frac{\mathbf{q}_t^{\top}\mathbf{R}_{-(t-i)}\mathbf{k}_i}{\sqrt{d}},
\label{eq:rope_attention}
\end{equation}
where $\mathbf{R}_t^{\top}\mathbf{R}_i=\mathbf{R}_{-(t-i)}$ depends only on the relative offset $t-i$, where $\Delta=t-i\geq0$ denotes the relative distance. Consequently, RoPE makes the QK score jointly dependent on the content representations $\mathbf{q}_t$ and $\mathbf{k}_i$ and their relative position.


Building on the relative-position-dependent scoring formulation above,
we denote task-relevant information far from the query as Distant Evidence $\mathcal{E}$ and nearby content not required for the prediction as Proximal Background $\mathcal{B}$. 
Importantly, $\mathcal{B}$ does not need to be adversarial, corrupted, or linguistically unnatural. It may consist of ordinary contextual or grammatical content that is coherent within the input but uninformative for the required prediction.    
\begin{remark}
Counterintuitively, distance does not determine task relevance: $\mathcal{E}$ can be distant but essential, whereas $\mathcal{B}$ can be nearby but uninformative. Nevertheless, tokens in both spans enter the same softmax normalization and therefore compete for attention weight.
\end{remark}

\subsection{Proximal Background Context Overshadows Distant Evidence}
\label{sec:2.2}


Building on the preceding definitions, we conduct two complementary experiments to examine the competition between distant evidence $\mathcal{E}$ and proximal background $\mathcal{B}$. The first determines whether pretrained LLMs fail to locate distant evidence or instead retrieve it but have its signal diluted by proximal background. The second compares moving the evidence closer with attenuating proximal background to identify which intervention more effectively restores evidence attention and answer confidence.

First, to distinguish whether the model fails to locate distant evidence or loses it under attention competition, we compare the layer-wise attention assigned to three regions: distant evidence, its neighboring tokens, and proximal background near the query. For each layer, we report the mean per-token attention density normalized by uniform attention, where a value of $1$ denotes the uniform baseline. These statistics are computed using Qwen3-8B \citep{bai2025longbench} on LongBench-v2 \citep{bai2025longbench}.

\begin{wrapfigure}{r}{0.4\textwidth}
    \centering
    \includegraphics[width=\linewidth]{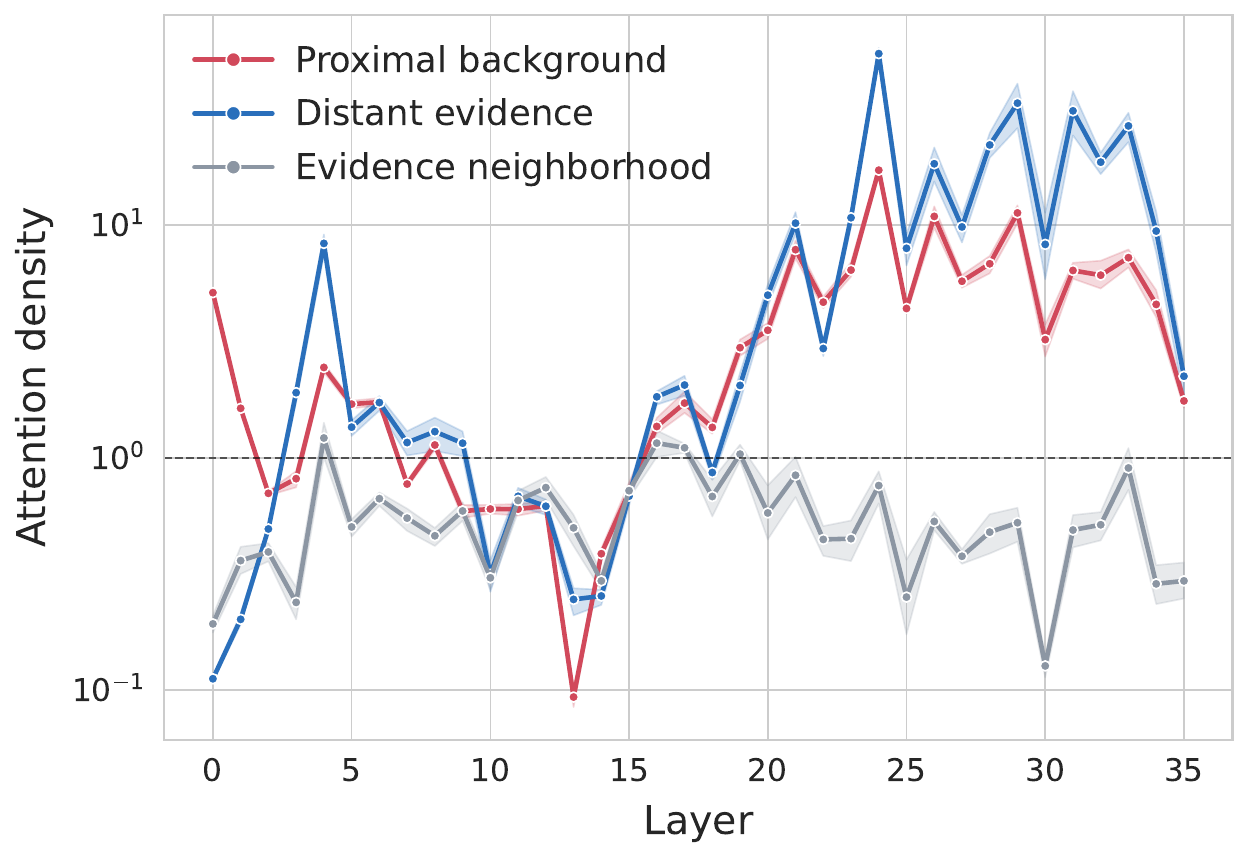}
    \captionof{figure}{\textbf{Distant Evidence is selectively retrieved but still diluted by cumulative competition from proximal background.} We report layer-wise attention density for distant evidence, its neighboring tokens, and proximal background near the query.}
    \label{fig:attention-density}
\end{wrapfigure}

As illustrated in Fig.~\ref{fig:attention-density}, the three regions are not consistently separated in the early and middle layers. From approximately layer 19 onward, however, attention to distant evidence increases sharply, whereas its neighborhood remains largely below the uniform baseline. This contrast shows that the model can precisely identify distant evidence among ordinary tokens at the same distant location. However, proximal background also receives substantial attention. Although its per-token attention is generally lower than that of distant evidence, its abundance produces sufficient cumulative attention mass to overshadow the evidence. The central issue is therefore not whether the model can locate distant evidence, but whether that evidence can withstand competition from abundant proximal background after being identified.

\begin{wrapfigure}{r}{0.4\textwidth}
    \centering
    \includegraphics[width=\linewidth]{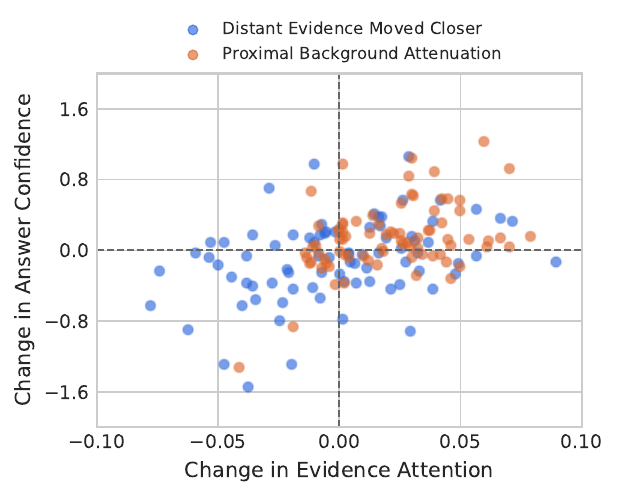}
    \captionof{figure}{\textbf{Proximal background limits evidence utilization in Qwen3-8B.} Points show paired changes in evidence attention and answer confidence. Blue moves evidence closer, while orange attenuates proximal background scores with positions fixed.}
    \label{fig:real_model}
\end{wrapfigure}

Subsequently, building on this observation, we examine whether evidence utilization in Qwen3-8B is constrained primarily by evidence distance or by competition from proximal background. Figure~\ref{fig:real_model} compares two interventions, moving distant evidence closer and attenuating proximal background, and reveals their distinct effects on evidence attention and answer confidence.
For each evidence-dependent question, we compare moving distant evidence closer with proximal background attenuation, where the latter suppresses QK scores of nearby task-irrelevant context without changing the input tokens or their positions. We report the resulting changes in Evidence Attention and correct-answer confidence relative to the original input.

Accordingly, as Fig.~\ref{fig:real_model} illustrates, the blue points span all four quadrants, indicating that moving distant evidence closer does not consistently improve the performance. In contrast, the orange points are concentrated toward increased evidence attention, with many also showing improved confidence. This divergence suggests that shortening the evidence distance does not remove competition from abundant proximal background. Attenuating proximal background instead directly weakens this competition, allowing more attention to be reassigned to relevant evidence. Therefore, compared with moving distant evidence closer, proximal background attenuation produces a more consistent improvement in evidence utilization.


Taken together, these two experiments reveal the following counter-intuitive finding:

\begin{finding}[\textbf{The Proximity Trap}]
Insufficient attention to distant evidence often arises less from distance itself. Instead, it is more from cumulative competition with abundant task-irrelevant proximal background context.
In this sense, proximal background acts as the Sirens' Song by drawing attention toward nearby but uninformative content and away from the distant evidence that matters.
\end{finding}

\subsection{Mechanism: Proximal Background Overshadowing}
\label{sec:2.3}

To better understand why task-relevant distant evidence can be overshadowed by task-irrelevant proximal background, we examine how relative position alters QK scores and how softmax converts these changes into attention competition.

Decomposing Eq.~\eqref{eq:rope_attention} into $d/2$ two-dimensional components, the QK score in the $r$-th component is

\begin{equation} 
\mathbf{q}_r^\top\mathbf{R}(-\Delta\omega_r)\mathbf{k}_r= a_r\cos(\Delta\omega_r) + b_r\sin(\Delta\omega_r), \label{eq:rope_pair_score} 
\end{equation}
where the content-dependent coefficients are defined as $a_r=q_{r,1}k_{r,1}+q_{r,2}k_{r,2}$ and $b_r=q_{r,1}k_{r,2}-q_{r,2}k_{r,1}$, and $\omega_r$ denotes the rotation frequency. Summing Eq.~\eqref{eq:rope_pair_score} across all components gives
\begin{equation}
s(\Delta)=\frac{1}{\sqrt{d}}\sum_{r=1}^{d/2}
\left[a_r\cos(\Delta\omega_r)+b_r\sin(\Delta\omega_r)\right].
\label{eq:rope_frequency_score}
\end{equation}

Because RoPE rotations are orthogonal, they preserve the query and key norms, namely $\lVert\widetilde{\mathbf{q}}_t\rVert_2=\lVert\mathbf{q}_t\rVert_2$ and $\lVert\widetilde{\mathbf{k}}_i\rVert_2=\lVert\mathbf{k}_i\rVert_2$. Thus, distance changes the directional match between the query and key across frequencies rather than reducing their vector lengths.

To isolate the positional effect in its simplest form, we consider a maximally aligned query--key pair with $\mathbf{q}_r=\mathbf{k}_r$. This serves only as an illustrative reference case and is not assumed to hold in pretrained LLMs. In this case, $b_r=0$ and $a_r=\lVert\mathbf{q}r\rVert_2^2$, reducing Eq.~\eqref{eq:rope_frequency_score} to
\begin{equation} 
s(\Delta)= \frac{1}{\sqrt{d}} \sum_{r=1}^{d/2} \lVert\mathbf{q}_r\rVert_2^2 \cos(\Delta\omega_r). \label{eq:aligned_rope_score} \end{equation}

For $|\Delta\omega_r|\ll1$, a second-order expansion gives:
\begin{equation}
s(\Delta)\approx s(0)-\frac{\Delta^2}{2\sqrt{d}}
\sum_{r=1}^{d/2}\lVert\mathbf{q}_r\rVert_2^2\omega_r^2.
\label{eq:local_rope_decay}
\end{equation}


Crucially, score attenuation alone does not determine the evidence attention. Consider one distant evidence token with QK score $s_{\mathcal E}$ and $M$ proximal background tokens, each with QK score $s_{\mathcal B}$. Their softmax competition gives
\begin{equation}
\alpha_{\mathcal E}=\frac{1}{1+M e^{s_{\mathcal B}-s_{\mathcal E}}},
\qquad
\alpha_{\mathcal E}<\frac{1}{2}\Longleftrightarrow
s_{\mathcal E}-s_{\mathcal B}<\log M.
\label{eq:background_competition}
\end{equation}

Accordingly, Eq.~\eqref{eq:background_competition} explains both experiments in Section~\ref{sec:2.2}. Fig.~\ref{fig:attention-density} shows that although the model attends to distant evidence, abundant proximal background can collectively overshadow it through $M$. Fig.~\ref{fig:real_model} further shows that attenuating proximal background restores evidence attention and answer confidence more consistently than moving the evidence closer. Thus, the primary obstacle is cumulative background competition rather than distance alone.

Together, these results distinguish vulnerability from failure: distance can weaken distant evidence, but insufficient evidence attention arises when abundant proximal background accumulates in the softmax denominator and collectively overshadows the evidence. 
This mechanism motivates an attention formulation that limits cumulative background competition while preserving access to distant evidence, leading to the LYRA formulation introduced next.

\section{LYRA for Distant Evidence}


The findings in Section~\ref{sec:2} reveal that distant evidence can be overshadowed by proximal background context, even when the latter contributes little task-relevant information. 
In this section, we introduce LYRA (\textbf{L}ong-context heav\textbf{Y}-tailed \textbf{R}elevance \textbf{A}lignment), a t-distributed directional matching method designed to preserve distant yet relevant evidence in Section~\ref{sec:3.1}, and analyze how it mitigates background-induced suppression in Section~\ref{sec:3.2}.

\subsection{LYRA: t-distributed QK Matching}
\label{sec:3.1}

As established in Section~\ref{sec:2}, Eq.\eqref{eq:background_competition} exhibits that even a moderate reduction in $s_{\mathcal E}$ can be exponentially amplified by softmax normalization, while the cumulative suppression from proximal background grows with $M$. A suitable mechanism should therefore preserve distant evidence under moderate score attenuation, distinguish evidence from background according to directional alignment with the query, and avoid indiscriminately increasing the weights of all distant tokens.

Motivated by these requirements, we introduce LYRA, a heavy-tailed directional matching mechanism that preserves attenuated yet relevant evidence without indiscriminately promoting distant context. Given the RoPE-transformed query \(\widetilde{\mathbf q}_t\) and key \(\widetilde{\mathbf k}_i\), we first measure their directional agreement using the normalized similarity
$
c_{t,i} = \frac{
\widetilde{\mathbf q}_t^{\top}\widetilde{\mathbf k}_i
}{
\lVert\widetilde{\mathbf q}_t\rVert_2
\lVert\widetilde{\mathbf k}_i\rVert_2
}
\in[-1,1].
$
We then reshape this similarity using the t-distributed transformation:

\begin{equation}
\phi_\kappa(c_{t,i})
=
\frac{1+c_{t,i}}{1+\kappa(1-c_{t,i})}-1,
\qquad \kappa\geq 0,
\label{eq:t-distributed}    
\end{equation}

where \(\kappa\) controls the angular concentration parameter
and \(\phi_0(c_{t,i})=c_{t,i}\) recovers the original cosine similarity. The resulting attention weight is defined as
$\alpha^{LYRA}
=\operatorname{softmax}_i
\left(\beta\phi_\kappa(c_{t,i})\right)$,
where $\beta>0$ is an inverse-temperature parameter. This formulation increases the contrast between highly aligned evidence and weakly aligned background without introducing an explicit position-dependent increase for distant tokens.

Operationally, LYRA modifies only the query-key scoring stage of attention. Instead of directly using the conventional QK score, it computes the directional similarity between the RoPE-transformed query and key and maps it to the score $\beta\phi_\kappa(c_{t,i})$ through Eq.~\eqref{eq:t-distributed}. RoPE, causal masking, softmax normalization, and value aggregation remain unchanged. LYRA can therefore be integrated into an existing attention layer by replacing its QK scoring function without modifying the remaining architecture.  The detailed implementation is provided in Appendix \ref{apdx:Implementation}.


\subsection{Preserving Distant Evidence under Background Competition}
\label{sec:3.2}

To characterize how LYRA locally reshapes directional similarity, we consider the derivative of the transformation in Eq.~\eqref{eq:t-distributed} and the critical similarity at which this derivative equals one:
\begin{equation}
\phi_\kappa'(c)
=
\frac{1+2\kappa}
{\left[1+\kappa(1-c)\right]^2}
>0,
\qquad
c_\kappa^*
=
1-\frac{2}{\sqrt{1+2\kappa}+1}.
\label{eq:LYRA_local_properties}
\end{equation}
The positive derivative shows that $\phi_\kappa$ is strictly increasing and therefore preserves the original similarity ordering. For $\kappa>0$, $c_\kappa^*$ divides the similarity domain into a compression regime with $c<c_\kappa^*$ and $\phi_\kappa'(c)<1$, and an amplification regime with $c>c_\kappa^*$ and $\phi_\kappa'(c)>1$. The transformation therefore reduces similarity differences in the former regime and enlarges them in the latter.

To extend the local characterization above to the finite difference between distant evidence and proximal background, we define the similarity-gap scaling factor for $c_{\mathcal B}\neq c_{\mathcal E}$ as
\begin{equation}
G_\kappa(c_{\mathcal B},c_{\mathcal E})
=
\frac{
\phi_\kappa(c_{\mathcal B})-\phi_\kappa(c_{\mathcal E})
}{
c_{\mathcal B}-c_{\mathcal E}
}=
\frac{1+2\kappa}
{
[1+\kappa(1-c_{\mathcal B})]
[1+\kappa(1-c_{\mathcal E})]
}.
\label{eq:LYRA_gap_scaling}
\end{equation}
Accordingly, the transformed gap satisfies $\phi_\kappa(c_{\mathcal B})-\phi_\kappa(c_{\mathcal E})=G_\kappa(c_{\mathcal B},c_{\mathcal E})(c_{\mathcal B}-c_{\mathcal E})$. As the secant slope of $\phi_\kappa$ between the two similarities, $G_\kappa<1$ indicates gap compression, whereas $G_\kappa>1$ indicates gap amplification. Compared with cosine-softmax under the same inverse temperature $\beta$, the relative change in the evidence-to-background attention ratio is
\begin{equation}
\frac{
\alpha_{\mathcal E}^{LYRA}/
\alpha_{\mathcal B}^{LYRA}
}{
\alpha_{\mathcal E}^{\cos}/
\alpha_{\mathcal B}^{\cos}
}
=
\exp\left[
\beta(1-G_\kappa)
(c_{\mathcal B}-c_{\mathcal E})
\right].
\label{eq:LYRA_relative_weight}
\end{equation}
The relative weight of evidence is therefore improved when $(1-G_\kappa)(c_{\mathcal B}-c_{\mathcal E})>0$. When proximal background has a higher similarity than distant evidence, $G_\kappa<1$ compresses the background advantage. When evidence retains a higher similarity, $G_\kappa>1$ enlarges the evidence advantage against cumulative background competition. Consequently, LYRA reduces the influence of weakly related background tokens while making strongly matched evidence more distinguishable, thereby directing attention toward task-relevant information.

Beyond competition from a single proximal background token, we extend the analysis to $M$ proximal background tokens, which does not change above improvement condition. The factor $M$ increases the cumulative competing mass, while $G_\kappa$ determines whether the score gap is adjusted in favor of distant evidence. Therefore, LYRA mitigates the Proximity Trap by narrowing a misleading background advantage or strengthening an existing evidence advantage before competition accumulates across background tokens. 

\subsection{ProxBench: Fine-Grained Proximal Perturbations for Long-Context}

Existing long-context benchmarks primarily evaluate whether models can retrieve and integrate information as the input length increases. However, they provide limited control over the proximal background context that competes with distant evidence. As a result, it remains difficult to determine whether a model succeeds by robustly identifying task-relevant evidence or fails because semantically confusable information near the query captures excessive attention. To address this limitation, we introduce \textsc{ProxBench}, a controlled benchmark designed to evaluate the robustness of long-context models against proximal background interference. \textsc{ProxBench} places sparse evidence at a distant position and introduces task-irrelevant but increasingly confusable background context near the query, thereby directly measuring the Proximity Trap under controlled conditions.

\textsc{ProxBench} organizes proximal perturbations into four progressively challenging levels. Level 1 introduces style-matched background that follows a similar syntactic structure but differs in entity, topic, and answer relation. Level 2 creates crossed bindings by placing the target entity and an access-code cue in the same sentence while explicitly associating the candidate value with another entity. Level 3 further increases ambiguity by mixing the same relation for different entities with different attributes of the target entity. Level 4 introduces fine-grained perturbations involving subtypes, attributes, semantic roles, and value formats, producing background context that closely resembles the required evidence while remaining logically irrelevant to the answer. This hierarchy enables a fine-grained assessment of how model performance changes as proximal background becomes increasingly difficult to distinguish from distant evidence. Further details on data construction, quality control, and dataset statistics are provided in Appendix~\ref{apdx:proxbench}.


\section{Experiments}


We organize our evaluation around three complementary aspects. First, in Section \ref{sec:4.1}, we evaluate LYRA on LongBench-v2 \citep{bai2025longbench} and  RULER \citep{hsieh2024ruler}, examining its ability to retrieve and utilize distant evidence under increasingly long inputs. Second, in Section \ref{sec:4.2}, we report results across the task categories of LongBench to assess whether the benefits of LYRA generalize beyond specific long-context settings. Third, in Section \ref{sec:4.3}, we evaluate LYRA on our proposed ProxBench, where progressively challenging perturbation levels are used to directly measure its robustness to different forms of proximal background interference. Additional ablation studies are provided in the Appendix \ref{sec:abl-exp} to examine the contribution and sensitivity of the key design in LYRA.

\textbf{Experimental Settings.}
We adopt Qwen3-8B \citep{yang2025qwen3}  as the base model and replace the standard attention mechanism only in its final Transformer block with LYRA, leaving the remaining architecture unchanged. We then fine-tune the model on LongAlign for one epoch, updating only the final Transformer block while freezing all preceding layers, token embeddings, and the language-modeling head. Training is performed in bfloat16 using AdamW, with a learning rate of $2\times 10^{-5}$, and a maximum sequence length of 16,384. We evaluate long-context retrieval and understanding on RULER and LongBench-v2, and assess robustness to proximity trap on ProxBench. The detailed derivation process is provided in Appendix \ref{apdx:Implementation}.

\begin{wraptable}{r}{0.61\textwidth}
\centering
\caption{\textbf{Evaluation results of different homogeneous methods on LongBench-v2 \citep{bai2025longbench},} following the evaluation setting of ProxyAttn \citep{wang2026proxy}. We compare our method against the baseline, MInference \citep{jiang2024minference}, FlexPrefill \citep{lai2025flex}, XAttention \citep{xu2025xatt}, ProxyAttn \citep{wang2026proxy}, S2O \citep{zhang2026s2o} and PBS-Attn \citep{wang2026sparser}. We report accuracy on the full evaluation set and across three context-length splits defined by word count: Short ($<$32K), Medium (32K--128K), and Long ($>$128K). The overall score is computed over the complete evaluation set. The best and second-best results in each column are highlighted in red and blue, respectively.}
\label{tab:longbench-v2}
\begin{adjustbox}{width=.6\textwidth}{
\begin{tabular}{@{}lccccc@{}}
\toprule
Methods       & Venue      & Short                                 & Medium                                & Long                                  & Overall                               \\ \midrule
Baseline      &            & 36.67                                 & 29.30                                 & 30.56                                 & 32.21                                 \\
Minference    & NeurIPS'24 & 40.60                                 & 26.50                                 & 28.70                                 & 32.02                                 \\
FlexPrefill   & ICLR'25    & 41.70                                 & {\color[HTML]{FF0000} 30.20}          & 27.80                                 & 33.80                                 \\
Xattention    & ICML'25    & 41.70                                 & 27.40                                 & {\color[HTML]{2F75B5} 32.40}          & 33.59                                 \\
ProxyAttn     & ICLR'26    & {\color[HTML]{2F75B5} 43.90}          & 28.40                                 & 27.80                                 & 33.82                                 \\
S2O      & ACL'26    & -                                     & -                                     & -                                     & 32.60          \\
PBS-Attn      & ICML'26    & -                                     & -                                     & -                                     & {\color[HTML]{2F75B5} 34.39}          \\ \midrule
\textbf{Ours} & \textbf{}  & {\color[HTML]{FF0000} \textbf{47.22}} & {\color[HTML]{2F75B5} \textbf{29.63}} & {\color[HTML]{FF0000} \textbf{33.33}} & {\color[HTML]{FF0000} \textbf{36.72}} \\ \bottomrule
\end{tabular}
}\end{adjustbox}
\end{wraptable}

\subsection{Main Results}
\label{sec:4.1}

First, we conduct experiments on LongBench-v2 \citep{bai2025longbench}, examining whether our LYRA improves long-context understanding across different input lengths. As exhibited in Table \ref{tab:longbench-v2},
LYRA achieves the best overall performance and consistently ranks among the strongest methods across all context-length splits. The leading performance on the Short split indicates that the improvement does not come at the cost of local context modeling, while the clear advantage on the Long split demonstrates a stronger ability to preserve and retrieve relevant evidence over extended contexts. This consistent behavior can be attributed to the combination of directional QK matching and the t-distributed transformation. Directional matching distinguishes semantically relevant evidence from weakly aligned background context, while the t-distributed transformation prevents relevant evidence from being excessively suppressed when its original QK score is weakened by distance. As a result, LYRA reduces the competition from proximal background tokens without indiscriminately increasing the scores of all distant tokens. The results therefore demonstrate that LYRA provides a robust balance between local context modeling and distant evidence utilization, supporting its effectiveness in mitigating the Proximity Trap.

\begin{table}[htbp]
\centering
\caption{\textbf{Evaluation results of different homogeneous methods on RULER \citep{hsieh2024ruler}}, following the evaluation setting of S2O \citep{zhang2026s2o}.
We compare our method against FlexPrefill \citep{lai2025flex}, XAttention \citep{xu2025xatt}, ProxyAttn \citep{wang2026proxy}, S2O \citep{zhang2026s2o}, and PBS-Attn \citep{wang2026sparser}. We report accuracy at five explicitly controlled context lengths, namely 8K, 16K, 32K, 64K, and 128K, providing a more fine-grained evaluation of length robustness. The average score is computed across these five context lengths. The best and second-best results in each column are highlighted in red and blue, respectively.}
\label{tab:ruler}
\begin{tabular}{@{}lccccccc@{}}
\toprule
              & Venue                            & 8K                                    & 16K                                   & 32K                                   & 64K                                   & 128K                                  & Avg.                                  \\ \midrule
FlexPrefill   & ICLR'25                          & 71.65                                 & 73.89                                 & 75.38  & 72.65                                 & 68.51                                 & 72.42                                                              \\
Xattention    & ICML'25                          & 85.63                                 & 82.25                                 & 81.60                                 & 73.18                                 & 69.91                                 & 78.51                                 \\
ProxyAttn     & ICLR'26                          & {\color[HTML]{2F75B5} 92.07}          & {\color[HTML]{2F75B5} 89.75}          & {\color[HTML]{2F75B5} 86.68}          & {\color[HTML]{2F75B5} 83.57}          & {\color[HTML]{2F75B5} 77.09}          & {\color[HTML]{2F75B5} 85.83}          \\
S2O           & ACL'26                           & 85.80                                 & 82.73                                 & 80.34                                 & 73.95                                 & 69.97                                 & 78.56                                 \\
PBS-Attn      & ICML'26                          & 85.56                                 & 79.34                                 & 80.95                                 & 70.70                                 & 67.90                                 & 76.89                                 \\ \midrule
\textbf{Ours} & {\color[HTML]{FF0000} \textbf{}} & {\color[HTML]{FF0000} \textbf{96.33}} & {\color[HTML]{FF0000} \textbf{94.19}} & {\color[HTML]{FF0000} \textbf{93.14}} & {\color[HTML]{FF0000} \textbf{85.39}} & {\color[HTML]{FF0000} \textbf{77.56}} & {\color[HTML]{FF0000} \textbf{89.32}} \\ \bottomrule
\end{tabular}
\end{table}

Moreover, we evaluate LYRA on RULER \citep{hsieh2024ruler} to provide a more fine-grained analysis of robustness to context length. Unlike LongBench-v2, which groups examples into broad context-length splits, RULER evaluates the same set of controlled tasks at explicitly specified input lengths, allowing the performance variation of each method to be examined as the context expands. 

As illustrated in Table \ref{tab:ruler}, LYRA achieves the best performance at every evaluated context length and establishes the strongest average result. The consistent advantage from 8K to 128K indicates that the effectiveness of LYRA is not restricted to a particular context range. In particular, LYRA delivers substantial improvements at intermediate and long context lengths while retaining the leading performance at 128K, where relevant evidence faces increasingly strong competition from background tokens. These results suggest that directional QK matching and the t-distributed transformation remain effective as the amount of competing context increases. Together with LongBench-v2, the consistent performance on RULER demonstrates that LYRA improves both practical long-context understanding and fine-grained length robustness.

\subsection{Generalization Across Diverse Tasks}
\label{sec:4.2}

Subsequently, we evaluate LYRA on LongBench \citep{bai2024longbench} to examine whether the improvements generalize across tasks with different evidence structures and reasoning requirements. In contrast to the preceding experiments, which focus on robustness across context lengths, this evaluation covers six task categories and therefore provides a broader assessment of task-level generalization. 

\begin{table}[htbp]
\centering
\caption{\textbf{Evaluation results of different homogeneous methods on LongBench \citep{bai2024longbench}.} We compare our method against MInference \citep{jiang2024minference}, FlexPrefill \citep{lai2025flex}, XAttention \citep{xu2025xatt}, PBS-Attn \citep{wang2026sparser}, Stem \citep{niu2026stem}, and Kascade \citep{deshmukh2026kascadepracticalsparseattention}. We report the official LongBench score across six task categories: single-document question answering (SQA), multi-document question answering (MQA), summarization, few-shot learning, code completion, and synthetic tasks. The best and second-best results in each column are highlighted in red and blue, respectively.
}
\label{tab:longbench}
\begin{tabular}{@{}lcccccccc@{}}
\toprule
Method        & Venue      & SQA                          & MQA                                   & Summ.                                 & Few-shot                              & Code                                  & Synthetic                    & Avg.                                  \\ \midrule
MInference    & NeurIPS'24 & {\color[HTML]{2F75B5} 46.92} & 37.11                                 & 16.53                                 & 34.15                                 & 2.98                                  & {\color[HTML]{2F75B5} 66.17} & 33.98                                 \\
FlexPrefill   & ICLR'25    & 46.08                        & 37.29                                 & 16.53                                 & 33.63                                 & 2.52                                  & 49.00                        & 30.84                                 \\
XAttention    & ICML'25    & 45.66                        & 37.77                                 & 16.66                                 & 36.20                                 & 2.02                                  & 64.67                        & 33.83                                 \\
PBS-Attn      & ICML'26    & {\color[HTML]{FF0000} 47.04} & 37.17                                 & 16.66                                 & 33.88                                 & 2.80                                  & {\color[HTML]{FF0000} 66.33} & 33.98                                 \\
STEM          & ICML'26    & -                            & 14.97                                 & 20.21                                 & 61.84                                 & 19.43                                 & 62.19                        & 35.73                                 \\
Kascade       & Arxiv'26   & 44.87                        & {\color[HTML]{2F75B5} 42.34}          & {\color[HTML]{2F75B5} 23.74}          & {\color[HTML]{2F75B5} 61.99}          & {\color[HTML]{FF0000} 62.71}          & 34.50                        & {\color[HTML]{2F75B5} 45.03}          \\ \midrule
\textbf{Ours} & \textbf{}  & \textbf{46.01}               & {\color[HTML]{FF0000} \textbf{43.35}} & {\color[HTML]{FF0000} \textbf{24.73}} & {\color[HTML]{FF0000} \textbf{62.17}} & {\color[HTML]{2F75B5} \textbf{58.41}} & \textbf{65.67}               & {\color[HTML]{FF0000} \textbf{50.06}} \\ \bottomrule
\end{tabular}
\end{table}

As presented in Table \ref{tab:longbench}, LYRA achieves the best average performance, with leading results on multi-document question answering, summarization, and few-shot learning, as well as competitive performance on the remaining tasks. The improvements are particularly clear for tasks that require relevant information to be identified and integrated from multiple locations. In multi-document question answering and summarization, evidence is often distributed across long contexts and competes with substantial background information, making these tasks especially susceptible to the Proximity Trap. Similarly, few-shot learning requires the model to identify useful demonstrations despite interference from other examples, while code completion often depends on distant definitions and dependencies surrounded by locally similar code. The strong performance across these tasks suggests that directional QK matching effectively distinguishes relevant evidence from weakly aligned background context, while the t-distributed transformation prevents useful information from being excessively suppressed by positional distance. Meanwhile, the competitive results on single-document question answering and synthetic tasks indicate that this mechanism does not compromise tasks dominated by more localized or explicit evidence. Overall, these results demonstrate that LYRA mitigates evidence competition across diverse task structures rather than benefiting only a specific context length or retrieval pattern.

\subsection{Alleviating the Proximity Trap on ProxBench}
\label{sec:4.3}

\begin{wrapfigure}{r}{0.6\textwidth}
    \centering
    \includegraphics[width=0.99\linewidth]{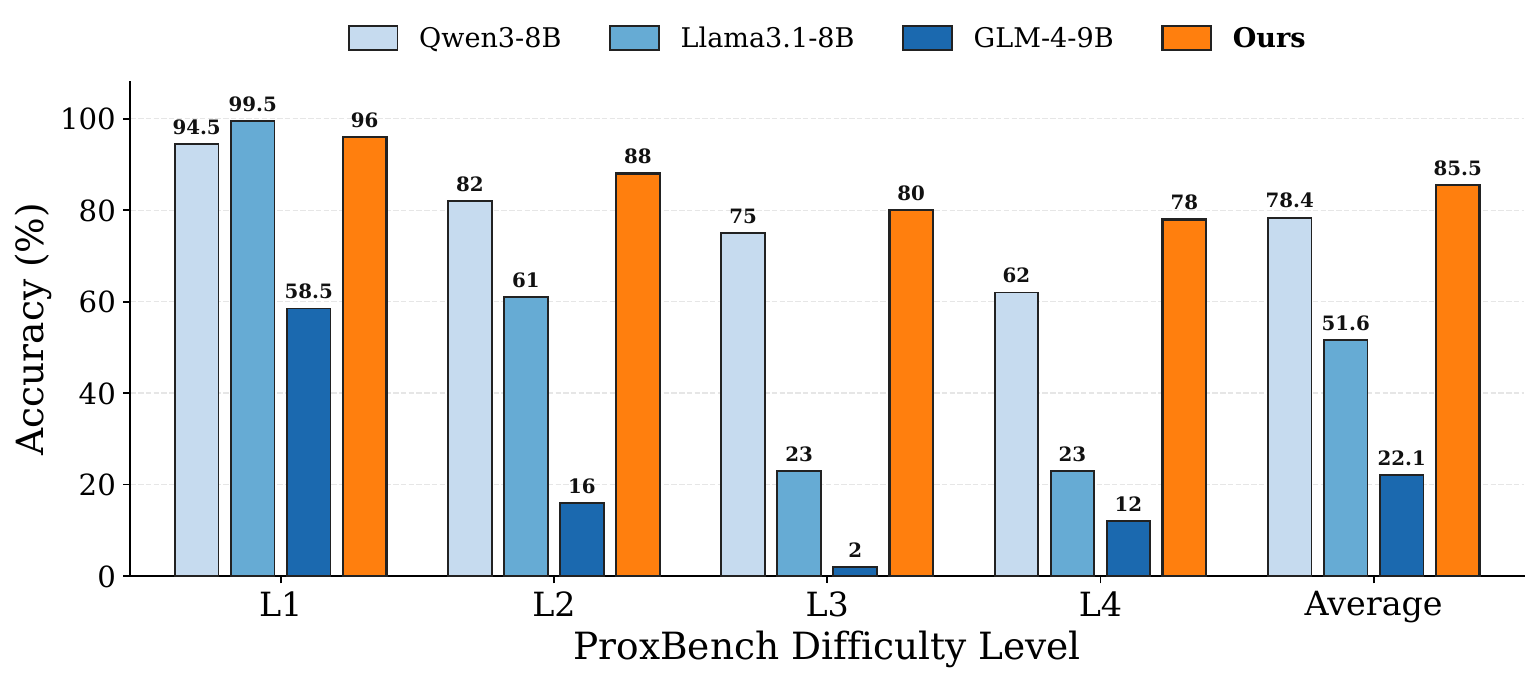}
    \caption{\textbf{Evaluation results on ProxBench across four progressively challenging proximal perturbation levels.} We compare Qwen3-8B \citep{yang2025qwen3}, Llama3.1-8B \citep{grattafiori2024llama}, GLM-4-9B \citep{glm2024chatglm}, and our LYRA model (8B) on Levels 1--4 and report accuracy (\%) at each level. The average score is computed over all four difficulty levels.}
    \label{fig:proxbench}
\end{wrapfigure}

Beyond standard long-context and generalization benchmarks, we evaluate LYRA on our proposed ProxBench to directly examine robustness to the Proximity Trap.
 The benchmark fixes the query and distant evidence while progressively increasing the semantic similarity and relational ambiguity of proximal background context. We compare LYRA with representative LLMs under the same evaluation protocol across four difficulty levels.

As shown in Figure~\ref{fig:proxbench}, LYRA achieves the best average performance and exhibits substantially greater robustness as the perturbation difficulty increases. Although Llama performs slightly better at the easiest level, LYRA consistently achieves the strongest results at all subsequent levels, with the advantage becoming particularly clear under the most fine-grained perturbations. In contrast, the performance of the baseline models declines sharply when the proximal background shares entities, relations, attributes, or value formats with the target evidence. The smaller performance degradation of LYRA indicates that the model does not rely primarily on positional proximity or surface-level similarity when identifying relevant information. Instead, directional QK matching captures fine-grained relevance between the query and evidence, while the t-distributed transformation prevents distant evidence from being overwhelmed by accumulated competition from proximal background tokens. These results provide direct evidence that LYRA mitigates the Proximity Trap by preserving task-relevant evidence even when nearby context is highly similar but logically irrelevant.

\section{Conclusion}

In this work, we identify the Proximity Trap, showing that distant evidence can be underutilized not only because of positional distance, but also due to cumulative competition from task-irrelevant proximal background. To address this issue, we introduce LYRA, a t-distributed directional matching mechanism that reshapes the context retrieval distribution to better preserve task-relevant evidence under background competition. Extensive experiments on LongBench-v2, RULER, LongBench, and ProxBench demonstrate consistent improvements across context lengths, task categories, and increasingly challenging proximal perturbations. Overall, our findings suggest that effective long-context modeling should consider not only how far relevant evidence lies from the query, but also what it must compete with along the way.

\bibliographystyle{plainnat}
\bibliography{references}

\appendix
\section*{Appendix}

\section{Related Works}

Our work is primarily related to two lines of research. The first develops long-context language models through context-window extension, long-context adaptation, and efficient attention computation. The second investigates positional biases and attention normalization, examining how the location of relevant information and competition among contextual tokens affect its influence on model predictions. We review these directions below and distinguish them from our focus on the competition between distant evidence and task-irrelevant proximal background.

\subsection{Long-Context Modeling}

Recent advances in long-context modeling have primarily focused on extending
the context window, adapting models to long sequences, and improving the
efficiency of long-context inference. One line of work extends the usable
context range through positional adaptation. YaRN rescales rotary positional
embeddings to support length extrapolation, while LongRoPE employs
non-uniform positional interpolation to accommodate substantially longer
sequences~\citep{peng2024yarn,pmlr-v235-ding24i}. More recently, RiPRA
incorporates semantic relevance into positional resource allocation, assigning
finer positional resolution to potentially useful content while compressing
less relevant regions~\citep{zhao2026adaptive}. These approaches substantially
expand the range over which contextual information remains representable, but
an extended context window does not necessarily ensure that distant evidence
is effectively incorporated into model predictions.

A complementary direction improves long-context capabilities through
adaptation and alignment. LongAlign constructs long-form instruction data and
develops corresponding training strategies to align language models with
long-context tasks~\citep{bai2024longalign}. Such training-based approaches
expose models to longer and more diverse dependencies, improving their ability
to operate over extended inputs. Nevertheless, their primary objective is to
establish general long-context competence, rather than to characterize how
distant evidence is utilized when it competes with abundant, task-irrelevant
context near the query.

Another substantial body of research seeks to reduce the computational and
memory costs of processing long sequences. StreamingLLM maintains stable
streaming inference by retaining attention sinks and a local context
window~\citep{xiao2024efficient}. Dynamic sparse-attention methods selectively
allocate computation to important regions: MInference exploits recurring
sparse attention patterns, FlexPrefill adapts sparse patterns and computational
budgets to individual inputs and attention heads, and XAttention estimates
block importance through antidiagonal scoring
~\citep{jiang2024minference,lai2025flex,xu2025xatt}. ProxyAttn further leverages
cross-head similarity to obtain more fine-grained block estimates with reduced
overhead~\citep{wang2026proxy}. Collectively, these methods make long-context
processing increasingly scalable by preserving selected attention regions
while avoiding exhaustive computation.

Despite this progress, existing studies largely concern whether distant
information remains positionally representable, computationally accessible,
or retained under efficient inference. Comparatively less attention has been
paid to whether a model can effectively use distant evidence after it
becomes accessible. In long inputs, distant evidence must compete with
abundant proximal background under normalized attention; consequently,
successful context extension or sparse selection alone does not guarantee that
the evidence receives sufficient influence over the final prediction. Our work
addresses this complementary problem by studying how proximal background
systematically suppresses the utilization of distant evidence and by improving
the allocation of attention under such contextual competition.

\subsection{Positional Bias and Attention Normalization}

A growing body of research has shown that the ability of language models to
use long contexts is strongly affected by the position of relevant
information. \citet{liu2024lost} reveal that models often perform well when
relevant information appears near the beginning or end of the input, but
struggle to use the same information when it is placed in intermediate
positions. Subsequent studies improve positional robustness through
long-context training. Position-agnostic decompositional training encourages
models to identify and integrate information across different context
positions~\citep{he2024never}, while structured packing organizes semantically
related documents into training sequences to improve the utilization of
distributed contextual information~\citep{staniszewski2025structured}. These
studies demonstrate that extending the context window alone does not guarantee
positionally robust information use and that targeted training can partially
alleviate context underutilization.

Complementary work investigates the mechanisms underlying such positional
preferences. Retrieval-head analyses identify a small subset of attention
heads that plays a disproportionate role in incorporating contextual evidence
into factual predictions~\citep{wu2025retrieval}. From a broader architectural
perspective, theoretical analyses attribute position bias to the joint effects
of causal masking, positional encoding, and information propagation across
layers~\citep{pmlr-v267-wu25ad}. Collectively, these findings suggest that the
influence of contextual evidence is shaped not only by its semantic relevance,
but also by where it appears and how attention propagates its information.
However, most existing analyses characterize the effect of evidence position
itself, leaving the role of the surrounding competing context comparatively
underexplored.

Another line of research examines the limitations of attention normalization.
Standard softmax couples all contextual tokens through a shared normalization
denominator, such that increasing the number of competing tokens can weaken
the selectivity of attention. Recent work improves its optimization properties
through input-dependent adjustment~\citep{zheng2025self}, while theoretical
analyses show that the ability of normalized attention to separate informative
from non-informative tokens deteriorates as the selected set or context grows
~\citep{mudarisov2025limitations}. These studies establish the general
limitations of softmax-based selection, but do not explicitly consider the
asymmetric competition between distant evidence and abundant proximal
background.

Our work connects positional bias with attention normalization through the
\emph{Proximity Trap}. Rather than treating the underutilization of distant
evidence solely as a consequence of its position, we show that it is jointly
driven by distance-dependent matching and the cumulative competition induced
by nearby, task-irrelevant context. Accordingly, LYRA strengthens attention to
task-relevant distant evidence while suppressing interference from weakly
relevant proximal background, without indiscriminately favoring all distant
tokens.

\section{ProxBench Details}
\label{apdx:proxbench}

\subsection{Task Formulation}

\textsc{ProxBench} evaluates whether a model can retrieve a target fact when the context contains records that partially overlap with the entity, relation, attribute, semantic role, or value format required by the query. Each instance samples a target entity $e$ from a predefined inventory of ten synthetic facility names and generates an access code $v$ in the format \texttt{AA-000}. The relevant evidence and query follow the templates
\begin{align}
    s^{+} &= \text{``The access code assigned to } e \text{ is } v \text{.''},\\
    q &= \text{``What is the access code assigned to } e \text{?''}.
\end{align}
The expected output is the code $v$ without additional explanation.

Each instance contains four proximal background records. These records are constructed to share selected surface or semantic cues with $s^{+}$ and $q$, while remaining logically insufficient for answering the query. In particular, the correct value never appears in the background, and no background record establishes the complete binding between the target entity, the queried access-code attribute, and the candidate value.

\subsection{Perturbation Levels}

The four levels control which components of the target fact are preserved in the background.  Let the target fact be represented as the tuple
\[
    (e,\; a_{\mathrm{access}},\; r_{\mathrm{assigned}},\; v),
\]
where $e$ is the target entity, $a_{\mathrm{access}}$ is the access-code attribute, $r_{\mathrm{assigned}}$ denotes the assignment relation and recipient role, and $v$ is the correct value.

\paragraph{Level 1: style-matched background.}
L1 controls for superficial similarity. Its records use the same declarative style and assignment-oriented syntax as the evidence, but change the entity, topic, and answer relation. Example attributes include \emph{maintenance schedule}, \emph{inspection timetable}, \emph{freight allocation}, and \emph{renovation plan}. Because these records contain neither the target entity nor a code-shaped candidate, they can be rejected using relatively coarse lexical and semantic cues.

\paragraph{Level 2: crossed bindings.}
L2 creates a local binding conflict by placing the target entity, an access-code phrase, another entity, and a candidate code in the same sentence. A typical template is:
\begin{quote}
\small
The [target entity] cross-reference lists [candidate code] as the access code assigned to [other entity], not to [target entity].
\end{quote}
Although most query-relevant tokens appear together, the semantic recipient of the candidate value is the other entity. Correct prediction therefore requires resolving the argument structure rather than copying the closest code-shaped span.

\paragraph{Level 3: mixed entity--relation interference.}
L3 alternates between two complementary perturbation types. The first preserves the full access-code relation but replaces the entity:
\begin{quote}
\small
The access code assigned to [other entity] is [candidate code].
\end{quote}
The second preserves the target entity and assignment relation but changes the queried attribute and value format:
\begin{quote}
\small
The emergency contact number assigned to [target entity] is [phone number].
\end{quote}
With four background records, each subtype appears twice. Consequently, the background contains both relation-matching candidates and entity-matching candidates, requiring the model to jointly resolve entity and attribute bindings.

\paragraph{Level 4: fine-grained binding characterization.}
L4 provides a finer diagnostic representation of the mixed binding condition. Each background record is assigned a semantic subtype describing the precise source of overlap, such as preserving the relation while replacing the entity or preserving the entity while replacing the attribute. The annotations additionally record entity, relation, attribute, and semantic-role overlap, as well as whether a candidate shares the expected code or phone-number format. This representation supports subtype-level error analysis and separates failures caused by entity confusion, attribute confusion, role reversal, and value-format attraction.

\subsection{Answer-Neutrality and Quality Control}

We apply the following automatic constraints during generation:

\begin{itemize}
    \item \textbf{No answer leakage.} The gold access code is prohibited from appearing in any background record.
    \item \textbf{Distinct candidate values.} Distractor codes are different from the gold answer and from other distractor codes in the same instance.
    \item \textbf{Entity separation.} Whenever a template requires another entity, it is sampled from the entity inventory after excluding the target entity.
    \item \textbf{Binding validity.} No background record asserts that a distractor value is the access code assigned to the target entity.
    \item \textbf{Exact background count.} Every instance contains four perturbation records, and all four spans must be recoverable after tokenization.
    \item \textbf{Length control.} Contexts are constructed under the target tokenizer so that the tokenized input plus the reserved generation budget exactly matches the requested sequence length.
    \item \textbf{Span verification.} Evidence, query, and background spans are located after serialization and re-tokenization. Instances that fail round-trip alignment are regenerated using a deterministic retry seed.
    \item \textbf{Reproducibility.} Instance-level seeds are derived from the global seed, context length, perturbation level, and example index using SHA-256. Dataset shards are additionally recorded with byte counts and SHA-256 checksums.
\end{itemize}

Each example also includes diagnostic metadata such as
\texttt{entity\_overlap},
\texttt{relation\_overlap},
\texttt{attribute\_overlap},
\texttt{role\_overlap},
\texttt{candidate\_value\_overlap},
\texttt{joint\_cue\_overlap},
\texttt{distractor\_value\_format}, and
\texttt{component\_strategies}.
These fields describe the intended interference mechanism and allow results to be decomposed by binding subtype.

\section{Implementation  and Computational Overhead}
\label{apdx:Implementation}

\paragraph{Implementation.}
For Qwen3-8B, we apply LYRA only to the final Transformer block
(block 35 under zero-based indexing). We optimize the complete final
block while freezing the preceding 35 blocks, token embeddings, final
normalization layer, and language-modeling head. This results in
approximately $193$M trainable parameters, corresponding to $2.36\%$
of the $8.19$B-parameter model. The head-specific LYRA parameters
themselves introduce only $2H=64$ additional scalars for $H=32$
query heads. Training is performed for one epoch in bfloat16 using
AdamW, with a maximum sequence length of $16{,}384$. We use a
per-device batch size of one on two devices, four gradient-accumulation
steps, a learning rate of $2\times10^{-5}$, a $3\%$ warmup ratio, zero
weight decay, and random seed 42.

\paragraph{Computational Overhead.}
Consider dense attention with batch size $B$, sequence length $n$,
$H$ query heads, and head dimension $d_h$. Standard attention requires
$\mathcal{O}(BHn^2d_h)$ arithmetic for query--key matching and value
aggregation. LYRA replaces the query--key product with a product between
normalized queries and keys and additionally performs
$\mathcal{O}(BHnd_h)$ normalization and
$\mathcal{O}(BHn^2)$ element-wise transformations. Its overall prefill
complexity therefore remains
$\mathcal{O}(BHn^2d_h)$, identical to dense attention asymptotically.
Autoregressive decoding likewise retains
$\mathcal{O}(BHnd_h)$ time per generated token, and LYRA does not alter
the $\mathcal{O}(BH_{\mathrm{kv}}nd_h)$ KV-cache size.

\begin{table}[htbp]
    \centering
    \caption{
        Analytical computation of Qwen3-8B and the additional
        FLOPs introduced by LYRA. The relative overhead remains below
        $0.04\%$ across all evaluated context lengths.
    }
    \label{tab:computational_overhead}
    \small
    \begin{tabular}{@{}lccc@{}}
        \toprule
        \textbf{Context}
        & Qwen3-8B (TFLOPs)
        & Additional LYRA (GFLOPs)
        & Relative Overhead\\
        \midrule
        8K   & 153.382     & 17.382   & 0.0113\% \\
        16K  & 385.929     & 69.122   & 0.0179\% \\
        32K  & 1,088.517   & 275.683  & 0.0253\% \\
        64K  & 3,443.670   & 1101    & 0.0320\% \\
        128K & 11,953.890  & 4401    & 0.0368\% \\
        \bottomrule
    \end{tabular}
\end{table}

Table~\ref{tab:computational_overhead} reports the analytical 
computation of Qwen3-8B and the additional FLOPs introduced by LYRA.
Across context lengths from 8K to 128K, the relative overhead increases
only from $0.0113\%$ to $0.0368\%$ and remains below $0.04\%$ in all
settings. This small overhead arises because LYRA modifies only the final
Transformer block and adds primarily element-wise normalization and
score-transformation operations, whose cost is minor relative to the
matrix multiplications in dense attention. Consequently, LYRA preserves
the asymptotic computational complexity of the base model while
introducing negligible additional arithmetic.

\section{Ablation Studies}
\label{sec:abl-exp}

\begin{table}[htbp]
\centering
\caption{\textbf{Ablation studies on different $\kappa$ on LongBench-v2 \citep{bai2025longbench}.} We vary $\kappa$, which controls the transformation strength of the t-distributed mapping, while keeping all other settings fixed. Accuracy is reported.}
\label{tab:abl-kappa}
\begin{tabular}{@{}ccccc@{}}
\toprule
$\kappa$ & Short & Medium & Long  & Overall \\ \midrule
2     & 41.11 & 27.91  & 31.48 & 33.40   \\
4     & 47.22 & 29.63  & 33.33 & 36.72   \\
8     & 44.44 & 28.84  & 30.63 & 34.81   \\
16    & 38.33 & 24.19  & 33.33 & 31.21   \\ \bottomrule
\end{tabular}
\end{table}

We conduct an ablation study to examine the effect of $\kappa$, which controls the strength of the t-distributed transformation. We vary $\kappa$, and evaluate each configuration across the context-length splits of LongBench-v2. As shown in Table~\ref{tab:abl-kappa}, the results exhibit a clear non-monotonic trend, with a moderate value of $\kappa$ achieving the strongest and most consistent performance across all context lengths. A smaller value provides insufficient separation between relevant evidence and weakly aligned background context, whereas an excessively large value applies overly aggressive score transformation and may suppress moderately aligned keys that still contain useful information. The consistent preference for the same intermediate setting across all splits indicates that $\kappa$ does not require length-specific tuning. These results confirm that effective mitigation of the Proximity Trap requires a balanced transformation that suppresses proximal background competition while preserving task-relevant evidence.

\end{document}